%% file: main.tex
\documentclass[runningheads]{llncs}
\usepackage{graphicx}
\graphicspath{ {./Tables/Images/} }
\usepackage{hyperref}

\usepackage{xcolor}
\usepackage{amsmath}

\usepackage{import}

\usepackage{caption}
\usepackage{subcaption}

\begin{document}
%
\title{Endo-Sim2Real: Consistency learning-based domain adaptation for instrument segmentation\thanks{Funded by the German Federal Ministry of Education and Research (BMBF) under the project COMPASS (grant no. - 16\,SV\,8019).}}
\titlerunning{Endo-Sim2Real}
%
\author{Manish Sahu\inst{1}
\and
Ronja Str{\"o}msd{\"o}rfer\inst{1}
\and
Anirban Mukhopadhyay\inst{2}
\and
Stefan Zachow\inst{1}
}
%
\authorrunning{Sahu et al.}
%
\institute{Zuse Institute Berlin (ZIB), Germany 
\and
Department of Computer Science, TU Darmstadt, Germany
}
\maketitle              
%


\begin{abstract}
Surgical tool segmentation in endoscopic videos is an important component of computer assisted interventions systems.
Recent success of image-based solutions using fully-supervised deep learning approaches can be attributed to the collection of big labeled datasets.
However, the annotation of a big dataset of real videos can be prohibitively expensive and time consuming.
Computer simulations could alleviate the manual labeling problem, however, models trained on simulated data do not generalize to real data.
This work proposes a consistency-based framework for joint learning of simulated and real (unlabeled) endoscopic data to bridge this performance generalization issue.
Empirical results on two data sets (15 videos of the Cholec80 and EndoVis'15 dataset) highlight the effectiveness of the proposed \emph{Endo-Sim2Real} method for instrument segmentation.
We compare the segmentation of the proposed approach with state-of-the-art solutions and show that our method improves segmentation both in terms of quality and quantity.
\end{abstract}

\section{Introduction}  \label{intro}

Segmentation of surgical tools in endoscopic videos is fundamental for realizing automation in computer assisted minimally invasive surgery or surgical workflow analysis.\footnote{EndoVis Sub-challenges - 2015, 2017, 2018 and 2019 (\href{https://endovis.grand-challenge.org}{URL})}
Recent research addresses the problem of surgical tool segmentation using fully-supervised deep learning.

One of the barriers for employing fully-supervised Deep Neural Networks (DNNs) is the creation of a sufficiently large ground truth data set. 
Even though the annotation process occupies valuable time of expert surgeons, it is essential for training fully-supervised DNNs.
While the research community is discussing effective ways for annotating surgical videos \footnote{SAGES Innovation Weekend - Surgical Video Annotation Conference 2020}, researchers are exploring possibilities beyond purely supervised approaches~\cite{fuentes2019easylabels,ross2018exploiting,pfeiffer2019generating}. 
A promising approach is utilizing virtual environments which provide a big synthetic dataset with clean labels for learning~\cite{pfeiffer2019generating}. 
However, DNNs trained purely on synthetic data (source domain) perform poorly when tested on real data (target domain) due to the domain/distribution shift problem~\cite{torralba2011unbiased}.
Domain adaptation~\cite{saenko2010adapting} is commonly employed to address this performance generalization issue.
In an annotation-free scenario, unsupervised domain adaptation (UDA) is applied since labels are not available for the target domain.
This can be achieved, for instance, by using a multi-step approach~\cite{pfeiffer2019generating},
where simulated images are first translated to realistic looking ones, 
before a DNN is trained with these pairs of translated images and their corresponding segmentation masks.
However, we argue that jointly learning from simulated and real data in an end-to-end fashion is important for adaptation of DNNs on both domains.

We propose an end-to-end, simulation to real-world domain adaptation approach for instrument segmentation, which does not require labeled (real) data for training a DNN (model).
At its core, there is an interplay between consistency learning for unlabeled real images and supervised learning for labeled simulated images.
The design and the loss functions of this approach are motivated by the key observation that the instrument shape is the most consistent feature between simulated and real images, while the texture and lighting may strongly vary.
Unlike Generative Adversarial Network (GAN) based approaches~\cite{pfeiffer2019generating} which mainly exploit style (texture, lighting) information~\cite{huang2018multimodal}, our proposed loss focuses on the consistency of tool shapes for better generalization.
Moreover, our approach, being end-to-end and non-adversarial by construction, is computationally more efficient (training time of hours vs days) in comparison to other multi-step and adversarial training approaches (\emph{I2I} and \emph{PAT}).

We demonstrate, by quantitative and qualitative evaluation of 15 videos of the \emph{Cholec80} dataset, that our proposed approach can either be used as an alternative or in conjunction with a GAN-based unsupervised domain adaptation method.
Moreover, we show generalization ability of our approach across an additional dataset (EndoVis~\cite{endovis2015}) and unseen instrument types.

\section{Related Work} \label{rel_work}

Research on instrument segmentation in minimally invasive procedures is dominated by fully-supervised learning using DNN architectures with: nested design~\cite{garcia2017toolnet}, recurrent layers~\cite{attia2017surgical}, dilated convolution~\cite{pakhomov2019deep} reusing pre-trained models~\cite{shvets2018automatic}, utilizing fusion modules~\cite{islam2019real} or attention modules~\cite{ni2019rasnet,ni2019raunet} to improve segmentation outcome. 
A comparison study on segmentation networks is, for instance, presented in~\cite{bodenstedt2018comparative}. 
Some researchers have focused on multi-task learning with landmark localisation~\cite{laina2017concurrent} or attention prediction~\cite{islam2019learning}, while  others use additional inputs, such as temporal priors from motion-flow~\cite{jin2019incorporating} or kinematic pose information~\cite{qin2019surgical}.
However, research in annotation efficient (semi-labeled) settings is limited. A weakly supervised learning approach is employed by~\cite{fuentes2019easylabels} utilizing weak scribbles labeling.
~\cite{ross2018exploiting} proposed  pre-training with auxiliary task (\emph{PAT}) on unlabelled data via self-supervised learning, followed by fine-tuning on a subset of labeled data in a semi-supervised learning setup.

The sole research work on unlabeled target with unsupervised learning is $I2I$~\cite{pfeiffer2019generating} which adopts a two-step strategy: first translating simulated images into real-looking laparoscopic images via a Cycle-GAN and then training these translated images with labels in a supervised manner.

Our work is in line with \emph{I2I} and focuses on the distribution shift problem via \emph{unsupervised domain adaptation}, where the DNN in not trained with labeled target data, unlike \emph{PAT}. 
However, in contrast to using a two-step approach~\cite{pfeiffer2019generating}, we employ an end-to-end consistency-based joint learning approach. 
An overview of our proposed approach in contrast to \emph{I2I} and \emph{PAT} is provided in Table~\ref{table:Comparison_Work}.

\begingroup
\begin{table}[t] 
	\centering
	\captionsetup{justification=centering}
	\caption{Comparison with current domain adaptation approaches.\newline ($L\rightarrow Labeled$ and $UL\rightarrow Unlabled$)}
	\label{table:Comparison_Work}
	\subimport{./Tables/}{Comparison_Work.tex}%
\end{table}
\endgroup

\begin{figure}[!b]
    \centering
    \includegraphics[width=\textwidth]{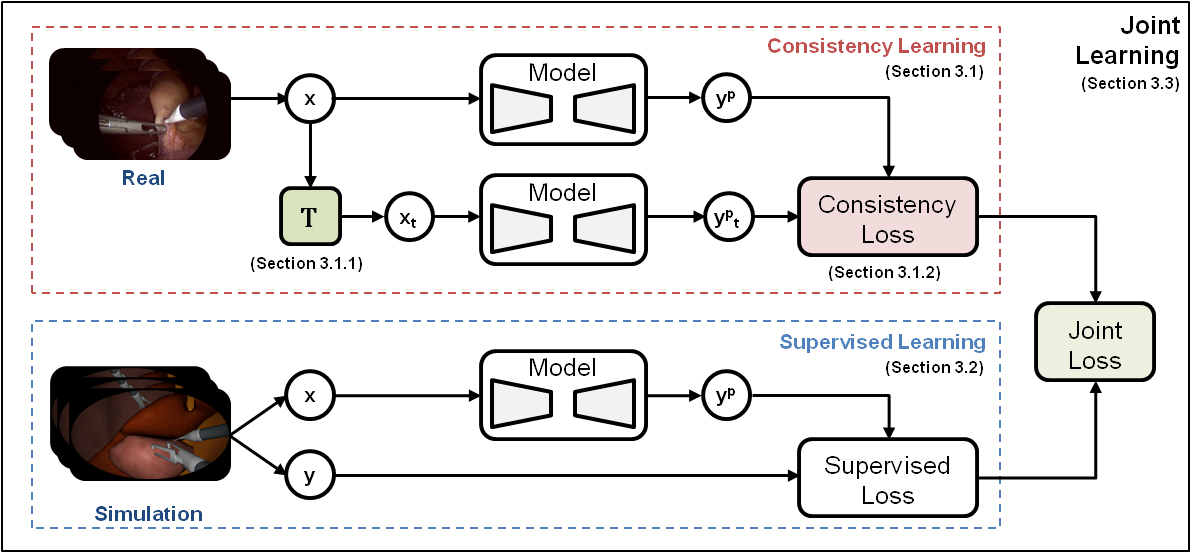}
    \caption{The proposed joint learning approach that involves supervised learning for simulated data and consistency learning for real unlabeled data.}
    \label{method:pipeline}
\end{figure}

\section{Method} \label{method}

Our proposed joint-learning framework (see Figure~\ref{method:pipeline}) is based on the principles of consistency learning.
It aims to bridge the domain gap between simulated and real data by aligning the DNN model on both domains.
For the instrument segmentation task we use TerNaus11~\cite{shvets2018automatic} as the DNN model as it is widely used and provides direct comparison with the previous work on domain adaptation (\emph{I2I}~\cite{pfeiffer2019generating}).
The rest of the core techniques of our proposed approach are described in this section.

\subsection{Consistency Learning} \label{consistency_learning}

The core idea of consistency learning (CL) is to learn meaningful representations by forcing a model ($f_{\theta}$) to produce a consistent output for an input which is \emph{perturbed in a semantics-preserving} manner~\cite{becker1992self,bachman2014learning,oliver2018realistic}.

\begin{equation}\label{Consistency Equation}
\min_{\theta}  D \{ f_{\theta}(x), f_{\theta}(x_{t}) \}
\end{equation}

In a general setting, given an input $x$ and its perturbed form $x_t$, the sensitivity of the model to the perturbation is minimized via a distance function $D(. , .)$.
The traditional CL setting~\cite{sajjadi2016regularization,laine2016temporal} uses mean-square-error (MSE) loss and commonly referred to as \emph{Pi-Model}.

\subsubsection{3.1.1. Perturbation (T)} \label{perturbation} 

A set of shape-preserving perturbations is necessary to learn the effective visual features in consistency learning. To facilitate learning, we implemented two data perturbation schemes: (1) \textit{pixel-intensity} (comprising random brightness and contrast shift, posterisation, solarisation, random gamma shift, random HSV color space shift, histogram equalization and contrast limited adaptive histogram equalization) and (2) \textit{pixel-corruption} (comprising Gaussian noise, motion blurring, image compression, dropout, random fog simulation and image embossing) and apply both of them in two ways:

\begin{itemize}
    \item \textbf{weak}: applying one of the pixel-intensity based perturbations
    \item \textbf{strong}: applying weak scheme followed by one of the pixel-corruption based perturbations
\end{itemize}

All perturbations are chosen in a stochastic manner. Please refer to supplementary material for more details regarding the perturbation schemes.

\subsubsection{3.1.2. Shape-based Consistency Loss (S-CL)}

In order to \emph{generalize} to a real domain, the model needs to focus on generic features of an instrument, like shape, and to \emph{adapt} to its visual appearance in an image.
Therefore, we propose to utilize a shape-based measure (\emph{Jaccard}) in conjunction with an entropy-based one (\emph{cross-entropy}).
In contrast to MSE, a shape-based loss formulation is invariant to the scale of the instrument as it depends on the overlap between prediction and ground-truth.
A combination of two losses is quite common in a fully-supervised setting~\cite{shvets2018automatic}.

\begin{equation}\label{Consistency loss Function}
\mathcal{L}_{cl} =  - \frac{1}{2} \left( \sum_{}^C y_t \log y_p + log{ \frac{|y_t\cap y_p|}{|y_t\cup y_p| - |y_t\cap y_p|}} \right)
\end{equation}

\subsection{Supervised Learning} \label{supervised_learning}

In order to learn effectively from the simulated image and its segmentation labels, we use the same combination of \emph{cross-entropy} and \emph{Jaccard} (see Eq. \ref{Consistency loss Function}) as the supervised loss function.

\subsection{Joint Learning}

Jointly training the model on both simulated and real domains is important for alignment on both dataset distributions and the main objective of domain adaptation.
Therefore, we employ two losses: supervised loss $\mathcal{L}_{sl}$ for labeled data and consistency loss $\mathcal{L}_{cl}$ for unlabeled data which are combined together through a joint loss, 

\begin{equation}\label{Joint loss Function}
\mathcal{L} =  \mathcal{L}_{sl} + \alpha * \mathcal{L}_{cl}
\end{equation}

The consistency loss is employed with a time-dependent weighting factor $\alpha$ in order to facilitate learning.
During training, the model updates the model predictions iteratively for the unlabeled data while training on both labeled and unlabeled data.
Thus, our joint loss minimization can be seen as propagating labels from labeled source data to unlabeled target data.

\section{Datasets and Experimental Setup} \label{experiments}

\subsection{Datasets}

\subsubsection{$Source$}

Source data is taken from~\cite{pfeiffer2019generating} which is acquired by translating rendered images (from 3-D laparoscopic simulations) to Cholec80 (images from in-vivo laparoscopic videos) style images \footnote{Please note that the rendered and real data sets are unpaired and highly unrelated.}.
The rendered dataset contains 20K images (2K images per patient) describing a random view of a laparoscopic scene with each tissue having a distinct texture and a presence of two tools (grasper and hook).
In this paper, we mention the specific datasets (from~\cite{pfeiffer2019generating}) by the following abbreviations:

\begin{itemize}
    \item \textbf{\textit{Sim}}: rendered dataset
    \item \textbf{\textit{SimRandom}} \textit{(SR)}: I2I translated images with five random styles
    \item \textbf{\textit{SimCholec}} \textit{(SC)}: I2I translated images with five Cholec80 image styles
\end{itemize}

\subsubsection{$Real^C$}

We have built an instrument segmentation dataset comprising of 15 videos of the Cholec80 dataset~\cite{Twinanda2016}. The segmentation dataset is prepared by acquiring frames at five frames per second, resulting in a total of 7170 images. These images are segmented for seven tools (grasper, hook, scissors, clipper, bipolar, irrigator and specimen bag).
In order to evaluate our approach, the dataset is divided into two parts: training (\textbf{\textit{$\mathbf{Real_{UL}^C}$}}) and testing (\textbf{\textit{$\mathbf{Real_{L}^C}$}}) dataset, containing 10 and 5 videos respectively.

\subsubsection{$Real^{EV}$}

We also evaluate on EndoVis'15 dataset for rigid instruments~\cite{endovis2015} containing 160 training (\textbf{\textit{$\mathbf{Real_{UL}^{EV}}$}}) and 140 testing (\textbf{\textit{$\mathbf{Real_{L}^{EV}}$}}) images, acquired from 6 in-vivo video recordings. Similar to $Real^C$ dataset, seven conventional laparoscopic instrument with challenging conditions like occlusion, smoke and bleeding are present in the dataset.

\subsection{Experimental Setup}

\subsubsection{Implementation}

We have built a standardised implementation where the model architecture, data processing and transformation, training procedure (optimizer, regularization, learning rate schedule, number of iterations etc.) are all shared in order to provide a direct comparison with other methods.
All the simulated input images and labels are first pre-processed with stochastically-varying circular masks to simulate the scene of real Cholec80 endoscopic images.
We are evaluating for (binary) instrument segmentation using dice score as the metric.
Also, we report all the results as an average of three training runs throughout our experiments. Our proposed framework is implemented in PyTorch and it takes four hours of training time on NVIDIA Quadro P6000 GPU.

\subsubsection{Hyper-parameters}

We use a batch size of 8 for 50 epochs and apply weight decay ($1e-6$) as standard regularization.
For consistency training, we linearly increase the weight of the unlabeled loss term $\alpha$ (see Eq. \ref{Joint loss Function}) from 0 to 1 over the training.
This linear increment of weight can be perceived as a warm up for the model, where it begins to understand the notion of instrument shape and linearly moves towards adapting to the real data distribution.


\section{Results and Discussion} \label{results}

\subsection{Quantitative Comparison}

\begingroup
\begin{table}[b] 
    \caption{Quantitative comparison using Dice score (std).}
	\label{table:Result_Compare}
	\subimport{./Tables/}{Comparisons.tex}
\end{table}
\endgroup

\subsubsection{Joint Learning on simulated data}

In this experiment we compared the performance of employing supervised learning on labeled simulated dataset ($Sim$) along with consistency learning on unlabeled real data in a traditional setting (Pi-Model) and our proposed approach for the instrument segmentation task on $Real_{L}^C$ and $Real_{L}^{EV}$ datasets.
The experimental results (Table~\ref{table:Result_Compare:CL}) highlight the usability of our joint learning approach as an alternative to the Cycle-GAN based \emph{I2I} framework.

\subsubsection{Joint Learning on translated data}

In this experiment we employed our approach on two translated datasets of \emph{I2I} ($SimRandom$ and $SimCholec$).
Our proposed approach is applied in conjunction with the $I2I$ approach and the results (Table~\ref{table:Result_Compare:I2I}) show the performance of joint learning against supervised learning on the translated images ($SimRandom$ and $SimCholec$).
The results indicate that our approach improves segmentation when used in conjunction with the GAN-based approach.

\subsection{Qualitative Analysis}

The images in Table~\ref{qualitative_analysis} depict the segmentation quality under challenging conditions.
Note that these conditions were absent in the simulated dataset.
Our proposed approach performs similar to $I2I$ when trained on the \emph{simulated dataset}. 
However, the segmentation quality improves and looks similar to that of the fully-supervised approach when trained on the \emph{translated dataset}.
Please refer to supplementary material for more details.

\begin{table}[t] 
    \centering
    \caption{Qualitative analysis with respect to artifacts \footnotesize{($X=Real_{L}^C$)}}
    \label{qualitative_analysis}
    \resizebox{1.0\columnwidth}{!}
    {\subimport{./Tables/}{Qualitative_Analysis.tex}}
\end{table}

\begingroup
\begin{table}[htbp]
    \caption{Performance comparison with respect to design choices. \footnotesize{($X=Real_{L}^C$)}}
	\label{ablation}
	\subimport{./Tables/}{Ablation.tex}
\end{table}
\endgroup

\subsection{Ablation Study on $Real^{C}$ dataset}

Since our consistency learning framework has multiple components, we study the effects of adding or removing a component to provide insights on reasons for the performance.

The experiments (Table~\ref{table:Result_Compare:SimDat}) show the performance on the \emph{simulated} dataset ($Sim$) with perturbation schemes (\emph{weak} and \emph{strong}) and consistency loss
functions (\emph{MSE} and \emph{S-CL}).
The substantial performance gap between \emph{Supervised Baseline} (supervised learning on source data only) and other approaches demonstrate the domain gap between simulated and real data and the effectiveness of employing our joint learning approach.
Strong perturbations in general perform better than weak perturbations, and our proposed approach outperforms Pi-Model.
The ablation (Table~\ref{table:Result_Compare:TransDat}) on the \emph{translated} dataset ($SimRandom$ and $SimCholec$) indicate that the consistency framework improves the overall performance of the DNN model from .75/.77 to .81/.80 on $SimRandom$ and $SimCholec$ datasets respectively. However, it also highlights that the choice of consistency losses (traditional and proposed loss) have low impact when the synthetic images are translated with $I2I$.
Please note that the same set of perturbations are applied as data augmentations for \emph{Supervised Baseline} (Table~\ref{table:Result_Compare:SimDat}) and \emph{I2I} (Table~\ref{table:Result_Compare:TransDat}) in order to provide direct and fair comparison.

\subsection{Generalization of tools from source to target} \label{results_generalization}

Our experiments highlight a generalization ability of the DNN model trained on labelled simulated data and unlabeled real data for the instrument segmentation task, considering that the simulated data contains only two conventional laparoscopic tools in contrast to the six tools (excluding specimen bag) in the real data. The results (Table~\ref{table:Result_Toolgeneralization}) highlight the strength of \emph{I2I} and our approach to generalize for unseen tools in the real domain.

\begingroup
\begin{table}[t] 
	\centering
	\caption{Comparison of tool generalization. \footnotesize{($X=Real_{L}^C$)}}
	\label{table:Result_Toolgeneralization}
	\subimport{./Tables/}{Tool_Generalization.tex}
\end{table}
\endgroup

\section{Conclusion} \label{disc}

We introduced an efficient and end-to-end joint learning approach for the challenging problem of domain shift between synthetic and real data.
Our proposed approach enforces the DNN to learn jointly from simulated and real data by employing a shape-focused consistency learning framework.
It also takes only four hours of training time in contrast to multiple days for the GAN-based approach~\cite{pfeiffer2019generating}.
Our proposed framework has been validated for the instrument segmentation task against the baseline (supervised learning on source data only), traditional CL approach (Pi-Model) and state-of-the-art $I2I$ approach. 
The results highlight that competitive performance (.75 vs .75/.77) can be achieved by using the proposed framework as an alternative. As a complementary method to the current GAN-based unsupervised domain adaption method, the performance further improves (to .81).
Finally, the generalization capabilities of our approach across two datasets and unseen instruments is highlighted to express the feasibility of utilizing virtual environments for instrument segmentation task.

In future, we plan to study the effects of perturbations in detail. 
Another avenue of investigation could be the direct learning of the perturbations from the data using Automatic Machine Learning~\cite{cubuk2019autoaugment}. 
Our method being flexible and unsupervised with respect to target data, can be further used for depth estimation by exploiting depth maps from the simulated virtual environments.
Since the community has not yet reached a consensus about how to efficiently label surgical videos, our end-to-end approach of joint learning provides a direction towards effective segmentation of surgical tools in the annotation scarce reality.



%
%
%
\bibliographystyle{splncs04}
\bibliography{references}

\end{document}


\title{Supplementary Material - Endo-Sim2Real}
\author{}
\institute{}
\maketitle
\titlerunning{Endo-Sim2Real}
\authorrunning{Sahu et al.}


\section{Additional Tables}  \label{tables}

\begingroup
\begin{table}[htbp] 
	\centering
	\setlength{\tabcolsep}{10pt} 
    \renewcommand{\arraystretch}{1.2} 
    \captionsetup{justification=centering}
	\caption{Performance comparison with respect to weight $\alpha$ for consistency loss in joint loss function,
	$\mathcal{L} =  \mathcal{L}_{sl} + \alpha * \mathcal{L}_{cl}$}
	\label{table:Result_CompareBaseline}
	\begin{tabular}{| l | c |}
	    	\hline
	    	Weight ($\alpha$) & $Sim\rightarrow Real_L^C$ \\
	    	\hline 
	    	\hline
	    	Constant & .69 \\ 
	    	Temporal & \textbf{.75} \\
	    	\hline
    \end{tabular}
\end{table}
\endgroup


\section{Additional Figures}  \label{figures}

\begin{figure}[htbp]
    \renewcommand{\figurename}{Figure}
    \renewcommand{\thefigure}{2}
    \includegraphics[width=\textwidth]{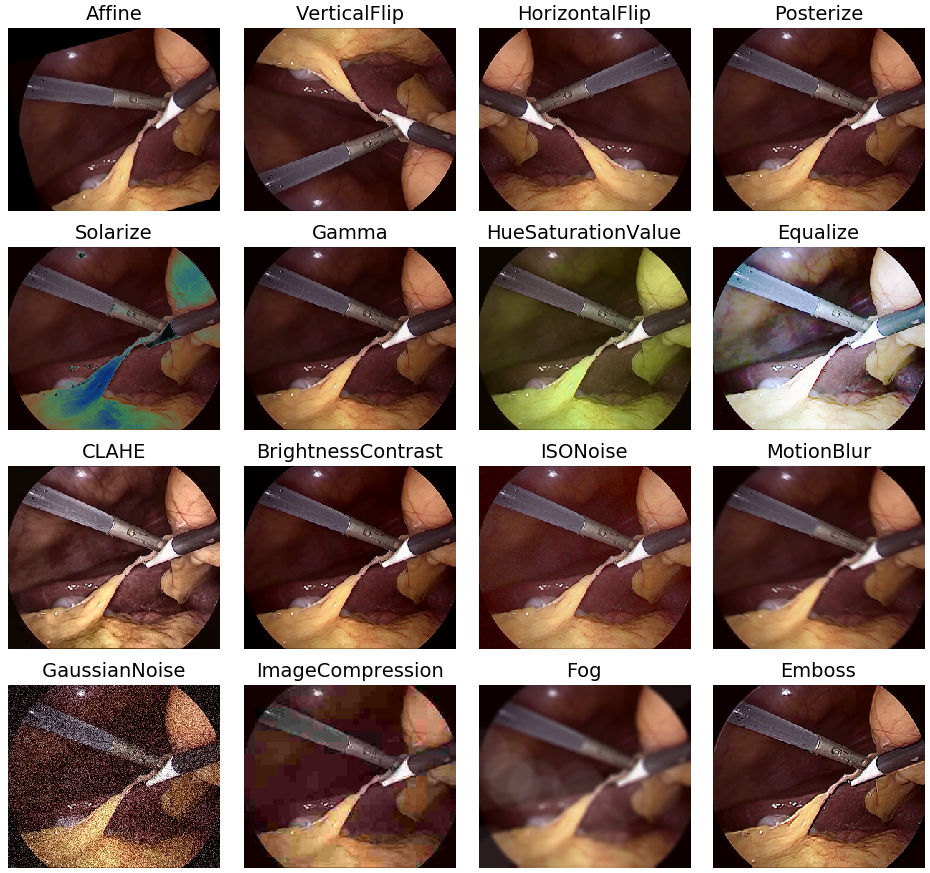}
    \small{\caption{All the images are augmented with affine and flip transformations before applying the perturbations schemes: \newline \textit{pixel-intensity} (comprising posterisation, solarisation, random gamma shift, random HSV shift, histogram equalization, contrast limited adaptive histogram equalization and random brightness and contrast shift) and \newline \textit{pixel-corruption} (comprising ISO noise, motion blurring, gaussian noise, image compression, random fog simulation and image embossing).}
    \label{figure:perturbations}}
\end{figure}


\begingroup
\begin{table}[htbp] 
    \centering
    \renewcommand{\tablename}{Figure} 
    \renewcommand{\thetable}{1} 
    \small{\caption{Qualitative analysis with respect to tool generalization \footnotesize{($X=Real_L^C$)}} \label{qualitative_analysis}}
    \subimport{SupplementaryTables/}{Clean.tex}
\end{table}
\endgroup

%% file: Tables/Comparison_Work.tex
\begin{tabular}{| c | c | c | c |}
	\hline
	Approach & Source (L$|$UL) & Target (L$|$UL) & Domain Adaptation (Steps) \\
	\hline
	\hline
	PAT ~\cite{ross2018exploiting} & Real (L) & Real (L) & Semi-supervised (Two-step) \\
	I2I ~\cite{pfeiffer2019generating} & Simulated (L) & Real (UL) & Unsupervised (Two-step) \\
	\hline 
	Ours & Simulated (L) & Real (UL) & Unsupervised (End-to-end) \\
	\hline
\end{tabular}

%% file: Tables/Comparisons.tex
\begingroup	
    \begin{subtable}{.5 \linewidth}
    \centering
    \caption{Comparison with Pi-Model} \label{table:Result_Compare:CL}
    \setlength{\tabcolsep}{3pt}
        \begin{tabular}{| l | c | c |}
	    	\hline
	    	Method & $Sim\rightarrow Real_{L}^C$ & $Sim\rightarrow Real_{L}^{EV}$ \\
	    	\hline 
	    	\hline
	    	Pi-Model    & .62 (.31)           & .54 (.32)\\ 
	    	Ours        & \textbf{.75} (.29)  & \textbf{.76} (.17)\\
	    	\hline
	    \end{tabular}
	\end{subtable} %
	\begin{subtable}{.5 \linewidth}
	\centering
    \caption{Comparison with I2I~\cite{pfeiffer2019generating}}
    \label{table:Result_Compare:I2I}
    \setlength{\tabcolsep}{3pt}
    	\begin{tabular}{| l | c | c |}
    		\hline
    		Method & $SR\rightarrow Real_{L}^C$ & $SC\rightarrow Real_{L}^C$ \\
    		\hline 
    		\hline
    		I2I & .75 (.30) & .77 (.29) \\ 
    		Ours & \textbf{.81} (.28) & \textbf{.80} (.28) \\ 
    		\hline
    	\end{tabular}
	\end{subtable} %
\endgroup

%% file: Tables/Qualitative_Analysis.tex
\begingroup
\begin{tabular}{|l|c c c c|}
    \hline
          & $Our_{Sim\rightarrow X}$ & $Our_{SimCholec\rightarrow X}$ & $I2I_{SimCholec\rightarrow X}$ & $FS_{Real_{UL}^C\rightarrow X}$ \\
        \hline
        \hline
        {Blood} & \parbox[c][1.5cm]{2.37cm}{\includegraphics[width=2.37cm, height=1.35cm]{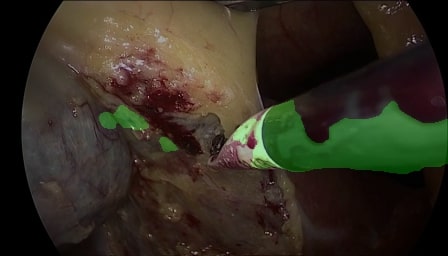}} & \parbox[c][1.5cm]{2.37cm}{\includegraphics[width=2.37cm, height=1.35cm]{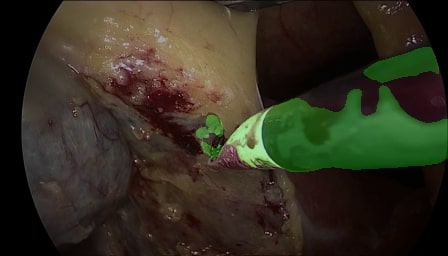}} & \parbox[c][1.5cm]{2.37cm}{\includegraphics[width=2.37cm, height=1.35cm]{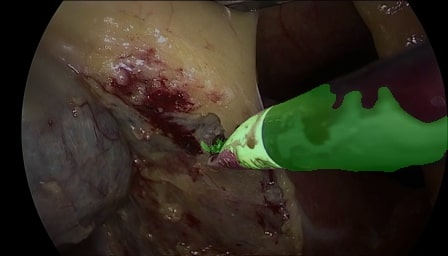}}  & \parbox[c][1.5cm]{2.37cm}{\includegraphics[width=2.37cm, height=1.35cm]{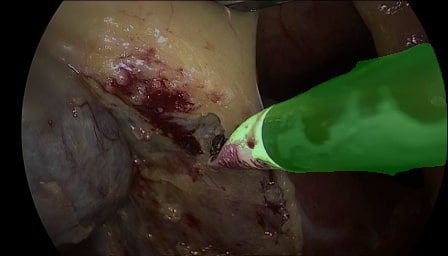}} \\
        \hline 
        \hline 
        {Motion} & \parbox[c][1.5cm]{2.37cm}{\includegraphics[width=2.37cm, height=1.35cm]{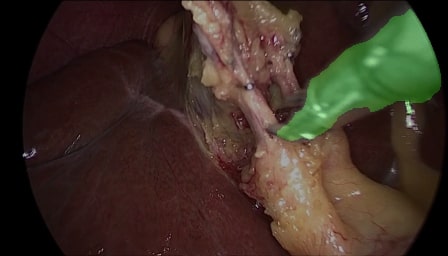}} & \parbox[c][1.5cm]{2.37cm}{\includegraphics[width=2.37cm, height=1.35cm]{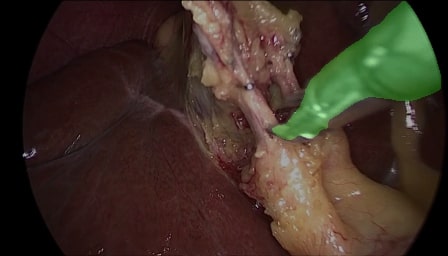}} & \parbox[c][1.5cm]{2.37cm}{\includegraphics[width=2.37cm, height=1.35cm]{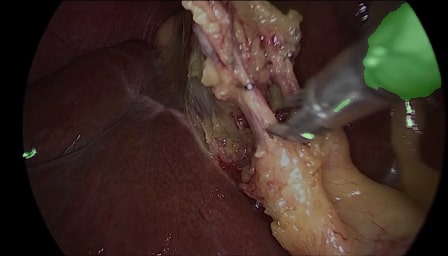}}  & \parbox[c][1.5cm]{2.37cm}{\includegraphics[width=2.37cm, height=1.35cm]{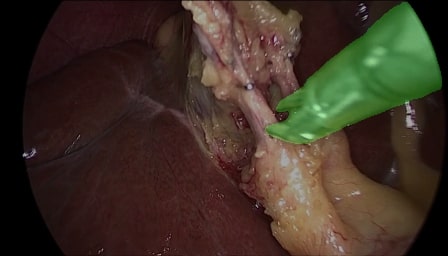}} \\
        \hline 
       {Smoke} & \parbox[c][1.5cm]{2.37cm}{\includegraphics[width=2.37cm, height=1.35cm]{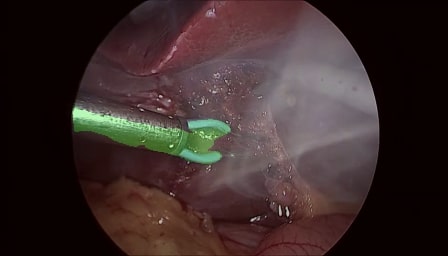}} & \parbox[c][1.5cm]{2.37cm}{\includegraphics[width=2.37cm, height=1.35cm]{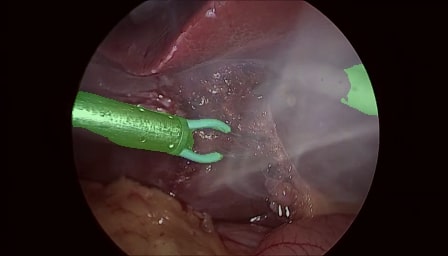}} & \parbox[c][1.5cm]{2.37cm}{\includegraphics[width=2.37cm, height=1.35cm]{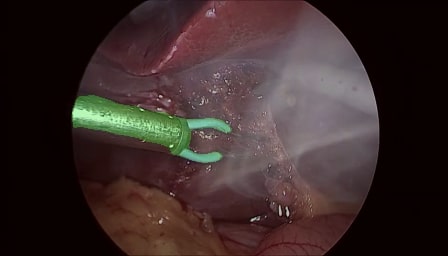}}  & \parbox[c][1.5cm]{2.37cm}{\includegraphics[width=2.37cm, height=1.35cm]{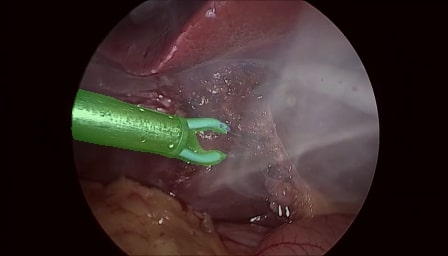}} \\
       \hline
\end{tabular}
\endgroup

%% file: Tables/Ablation.tex
\begingroup    
    \begin{subtable}{.5\linewidth}
        \centering
        \caption{Ablation on simulated dataset}
	    \label{table:Result_Compare:SimDat}
    	\setlength{\tabcolsep}{3pt}
        \begin{tabular}{| l | c |}
		\hline
		Loss + T & $Our_{Sim\rightarrow X}$  \\
		\hline 
		\hline
		Supervised Baseline & .30 \\
		MSE $+$ Weak        & .56 \\ 
		S-CL $+$  Weak      & .69 \\ 
		MSE $+$ Strong      & .62 \\ 
		S-CL $+$ Strong     & \textbf{.75}\\ 
		\hline
		\end{tabular}
	\end{subtable}%
    \begin{subtable}{.5\linewidth}
	\centering
	\caption{Ablation on translated dataset}
	\label{table:Result_Compare:TransDat}
	\setlength{\tabcolsep}{3pt}
	    \begin{tabular}{| l | c | c |}
		    \hline
		    Loss + T & ${SR\rightarrow X}$ & ${SC\rightarrow X}$ \\
    		\hline 
    		\hline
	    	I2I                 & .75 & .77  \\
	    	MSE $+$ Weak   & .72 & .76  \\
		    S-CL $+$  Weak        & .73 & .77  \\
		    MSE $+$ Strong & .79 & .81  \\
		    S-CL $+$ Strong       & \textbf{.81} & \textbf{.80}  \\
	    	\hline
	    \end{tabular}
    \end{subtable}%
\endgroup

%% file: Tables/Tool_Generalization.tex
\begin{tabular}{| l | c | c | c || c |}
	\hline
	Tools & $Our_{Sim\rightarrow X}$ & $Our_{SimCholec\rightarrow X}$ & $I2I_{SimCholec\rightarrow X}$ & $FS_{Real_{UL}^C\rightarrow X}$ \\
	\hline 
	\hline
	Grasper & .72 & \textbf{.81} & .75 & \textbf{.90} \\ 
	Hook & .82 & \textbf{.88} & .84 & \textbf{.93}  \\ 
	Scissors & .70 & \textbf{.76} & .72 & \textbf{.91} \\ 
	Clipper & .75 & \textbf{.85} & .77 & \textbf{.89} \\ 
	Bipolar & .70 & \textbf{.80} & .72 & \textbf{.89}  \\ 
	Irrigator & .74 & \textbf{.82} & .78 & \textbf{.88} \\
	\hline
\end{tabular}

%% file: SupplementaryTables/Clean.tex

\begin{tabular}{|l|c c c c|}
    \hline
          & $Our_{Sim\rightarrow X}$ & $Our_{SimCholec\rightarrow X}$ & $I2I_{SimCholec\rightarrow X}$ & $FS_{Real_{UL}\rightarrow X}$ \\
        \hline
        \hline
        {\tiny{Grasper \&  Hook }} & \parbox[c][1.5cm]{2.37cm}{\includegraphics[width=2.37cm, height=1.35cm]{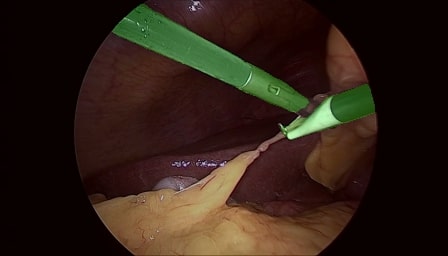}} & \parbox[c][1.5cm]{2.37cm}{\includegraphics[width=2.37cm, height=1.35cm]{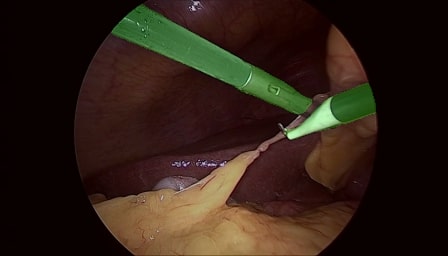}} & \parbox[c][1.5cm]{2.37cm}{\includegraphics[width=2.37cm, height=1.35cm]{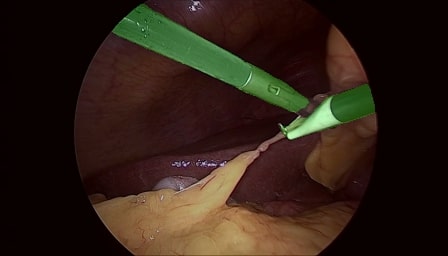}} & \parbox[c][1.5cm]{2.37cm}{\includegraphics[width=2.37cm, height=1.35cm]{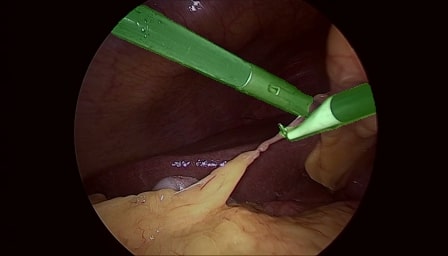}} \\
        \hline 
        {\tiny{Bipolar}} & \parbox[c][1.5cm]{2.37cm}{\includegraphics[width=2.37cm, height=1.35cm]{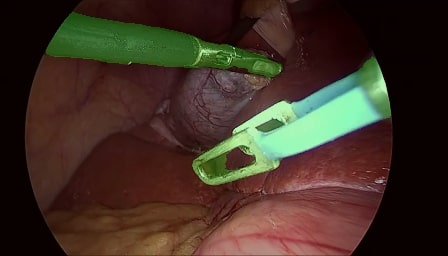}} & \parbox[c][1.5cm]{2.37cm}{\includegraphics[width=2.37cm, height=1.35cm]{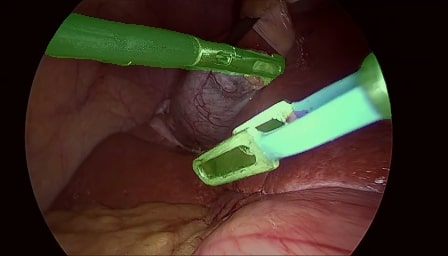}} & \parbox[c][1.5cm]{2.37cm}{\includegraphics[width=2.37cm, height=1.35cm]{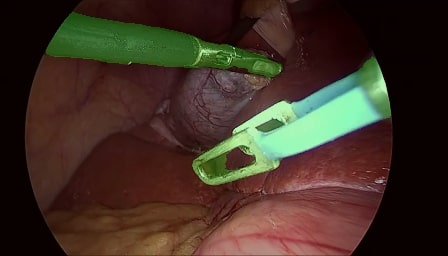}} & \parbox[c][1.5cm]{2.37cm}{\includegraphics[width=2.37cm, height=1.35cm]{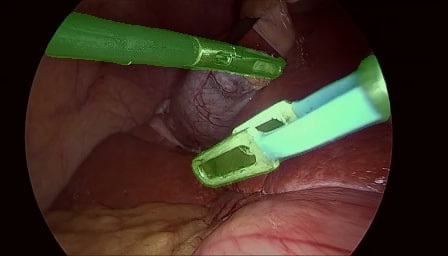}} \\
        \hline 
        {\tiny{Clipper}} & \parbox[c][1.5cm]{2.37cm}{\includegraphics[width=2.37cm, height=1.35cm]{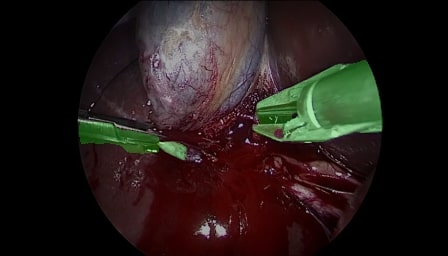}} & \parbox[c][1.5cm]{2.37cm}{\includegraphics[width=2.37cm, height=1.35cm]{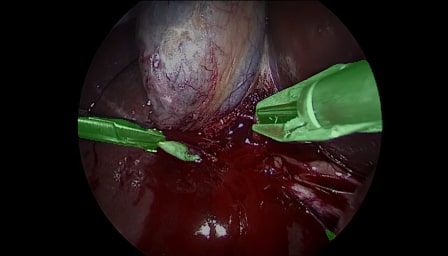}} & \parbox[c][1.5cm]{2.37cm}{\includegraphics[width=2.37cm, height=1.35cm]{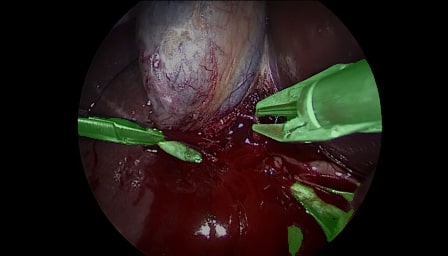}} & \parbox[c][1.5cm]{2.37cm}{\includegraphics[width=2.37cm, height=1.35cm]{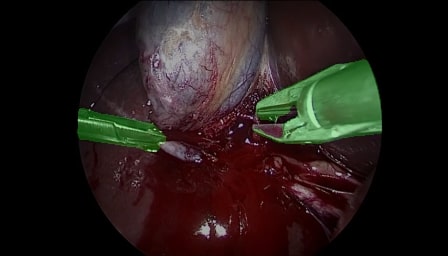}} \\
        \hline 
        {\tiny{Scissors}} & \parbox[c][1.5cm]{2.37cm}{\includegraphics[width=2.37cm, height=1.35cm]{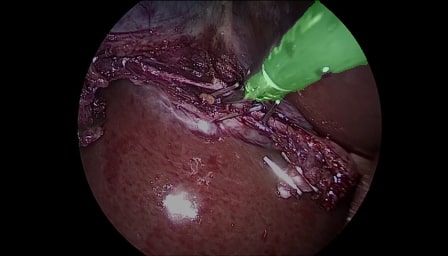}} & \parbox[c][1.5cm]{2.37cm}{\includegraphics[width=2.37cm, height=1.35cm]{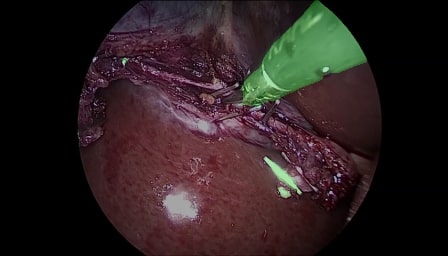}} & \parbox[c][1.5cm]{2.37cm}{\includegraphics[width=2.37cm, height=1.35cm]{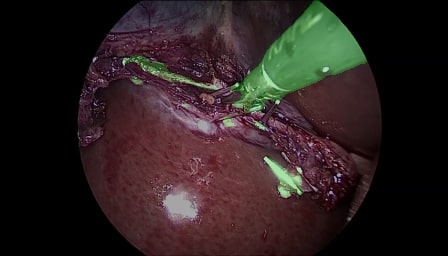}} & \parbox[c][1.5cm]{2.37cm}{\includegraphics[width=2.37cm, height=1.35cm]{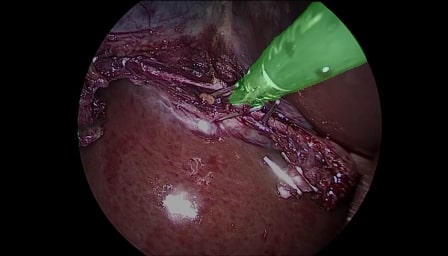}} \\
        \hline 
        {\tiny{Irrigator}} & \parbox[c][1.5cm]{2.37cm}{\includegraphics[width=2.37cm, height=1.35cm]{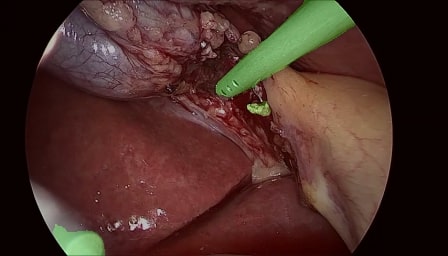}} & \parbox[c][1.5cm]{2.37cm}{\includegraphics[width=2.37cm, height=1.35cm]{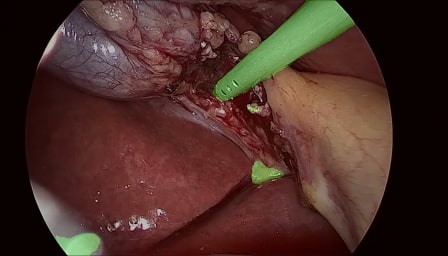}} & \parbox[c][1.5cm]{2.37cm}{\includegraphics[width=2.37cm, height=1.35cm]{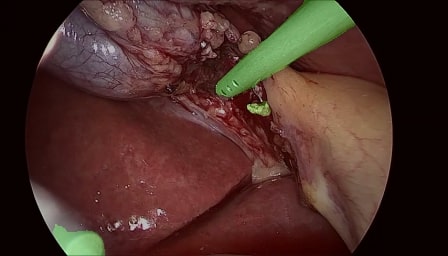}} & \parbox[c][1.5cm]{2.37cm}{\includegraphics[width=2.37cm, height=1.35cm]{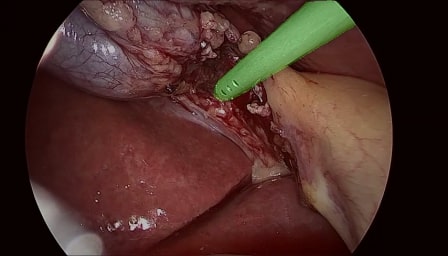}} \\
        \hline
\end{tabular}
